\def\year{2020}\relax
\documentclass[letterpaper]{article} 
\usepackage{aaai20}  
\usepackage{times}  
\usepackage{helvet}  
\usepackage{courier}  
\usepackage[hyphens]{url}  
\usepackage{graphicx}  
\usepackage{subcaption}
\usepackage{amsfonts}
\usepackage{amssymb}
\usepackage{xspace}
\usepackage{booktabs}
\usepackage{amsmath,bm}
\usepackage{dsfont}
\usepackage{multirow}
\usepackage[autostyle, english = american]{csquotes}
\usepackage[ruled,linesnumbered]{algorithm2e}

\MakeOuterQuote{"}
\usepackage{ntheorem,lipsum} 
\theorembodyfont{\upshape} 
\newtheorem{definition}{Definition}
\newtheorem{problem}{Problem}
\usepackage{enumitem}
\usepackage{xcolor}
\usepackage{nameref}
\newcommand{\mname}{\texttt{CONAN}\xspace}

\usepackage[colorlinks=true, allcolors=blue]{hyperref}

\begin{document}
%

\title{\mname: Complementary Pattern Augmentation  for Rare Disease Detection}
\author{Authors
}

\author{
\Large \textbf{Limeng Cui\textsuperscript{\rm 1,2}, Siddharth Biswal\textsuperscript{\rm 1,3}, Lucas M. Glass\textsuperscript{\rm 1}, Greg Lever\textsuperscript{\rm 1}, Jimeng Sun\textsuperscript{\rm 3}}, Cao Xiao\textsuperscript{\rm 1} \\
\textsuperscript{\rm 1} Analytic Center of Excellence, IQVIA, Cambridge, MA, USA \\
\textsuperscript{\rm 2} College of Information Sciences and Technology, The Pennsylvania State University, PA, USA \\ 
\textsuperscript{\rm 3}  College of Computing, Georgia Institute of Technology, Atlanta, GA, USA 
}

\maketitle

\begin{abstract}
Rare diseases affect hundreds of millions of people worldwide but are hard to detect since they have extremely low prevalence rates (varying from 1/1,000 to 1/200,000 patients) and are massively underdiagnosed. How do we reliably detect rare diseases with such low prevalence rates? How to further leverage patients with possibly uncertain diagnosis to improve detection?
In this paper, we propose a \underline{Co}mplementary patter\underline{n} \underline{A}ugmentatio\underline{n} (\mname) framework for rare disease detection. \mname combines ideas from both adversarial training and max-margin classification. It first learns self-attentive and hierarchical embedding for patient pattern characterization. Then, we develop a complementary generative adversarial networks (GAN) model to generate candidate positive and negative samples from the uncertain patients by encouraging a max-margin between classes. In addition, \mname has a disease detector that serves as the discriminator during the adversarial training for identifying rare diseases. We evaluated \mname on two disease detection tasks. For low prevalence inflammatory bowel disease (IBD) detection, \mname achieved $.96$ precision recall area under the curve (PR-AUC) and $50.1\%$ relative improvement over the best baseline. For rare disease idiopathic pulmonary fibrosis (IPF) detection, \mname achieves $.22$ PR-AUC with $41.3\%$ relative improvement over the best baseline.

\end{abstract}

\section{Introduction}
\label{sec:intro}

There are more than 7,000 of diseases that are individually rare but collectively common. These rare diseases affect 350 million people worldwide and incur a huge loss in quality of life and large financial cost~\cite{Vickers19}.
As these diseases are rare individually, initial misdiagnosis is common. On average it can take more than seven years for rare disease patients to receive the accurate diagnosis with the help of 8 physicians \cite{shire2013rare}. Thus, it is important to detect and intervene with the rare disease before it becomes life-threatening and consumes excessive medical resources. In recent years, the availability of massive electronic health records (EHR) data enables the training of deep learning models for accurate predictive health~\cite{doi:10.1093/jamia/ocy068}. However, current success mainly focuses on common chronic diseases such as Parkinson's disease progression modeling~\cite{baytas2017patient,doi:10.1137/1.9781611974973.23} and heart failure prediction \cite{choi2018mime}, deep learning models for rare disease prediction are lacking.


Two key challenges are presented for rare disease detection. First, the \textbf{low prevalence rates} of rare diseases limit the number of positive subjects in the training data (i.e., patients with a confirmed diagnosis of the rare disease). Thus the disease patterns are hard to extract. Second, there exist many \textbf{patients with uncertain diagnosis} due to the long period needed for rare diseases to be correctly diagnosed. The existence of large number of such uncertain patients can potentially help the disease detector perform better, as they are inherently close to positive patients who has confirmed diagnosis of the rare disease.

Related setting can be found in Positive-Unlabeled (PU) learning, which assumes the unlabeled data can contain both positive and negative examples~\cite{bekker2018learning}. Existing PU learning methods 
~\cite{elkan2008learning,kiryo2017positive} 
often identify reliable negative examples and then learn based on the labeled positives and reliable negatives ~\cite{liu2003building}. 
However, it is difficult to apply existing PU learning methods for the rare disease detection problem as the key difficulty here is to distinguish positive patients from negative ones with similar conditions. The reliable negative examples (e.g., healthy individuals and patients with similar diseases) will yield a more relaxed classification hyperplane and thus generate many false positive cases due to the low prevalence rates of rare diseases. 

To mitigate the aforementioned problem, researchers adopt the generative adversarial networks (GANs)~\cite{NIPS2014_5423} to augment the minor class and balance the distribution~\cite{mclachlan2016using,choi2017generating}. However, several challenges are still present in GAN based methods:
\begin{enumerate}
    \item They often focus on generating raw patient data, which is an extremely difficult problem on its own. The resulting synthetic data can easily be non-realistic and thus not useful for disease detection. 
    \item They often try to augment data based on the rare class only. However, due to the low prevalence rate, rare class (rare disease patients) are insufficient to provide robust embedding that supports effective data augmentation.
    \item Many GAN based methods often generate positive samples from Gaussian noises and apply a discriminator to distinguish real from fake data, which is not targeted toward rare disease detection.
\end{enumerate}

To tackle these challenges, we propose {\it pattern augmentation} that can better preserve and enrich crucial patterns of the target disease. We also recognize that among negative subjects, there exist ``borderline'' cases
that have uncertain diagnoses and potentially have the risk of rare diseases. For example, a rare disease idiopathic pulmonary fibrosis (IPF) shares compatible clinical, radiological, and pathological findings with a common chronic disease hypersensitive pneumonitis~\cite{cordeiro2013clinical}. Without any further investigation, no definite diagnosis could be made, leaving many patients to be \textit{uncertain}.

Based on the above observations, we propose the \underline{Co}mplementary patter\underline{n} \underline{A}ugmentatio\underline{n} (\mname) framework which combines the idea of adversarial training and max margin classification for accurate rare disease detection.
\mname is enabled by the following technical contributions:

\begin{enumerate}
    \item \textbf{Self-attentive and hierarchical embedding for better patient pattern characterization}. \mname constructs an end-to-end hierarchical (visit- and patient-level) embedding model with two levels of self-attention to embed the raw patient EHR into latent pattern vectors. The resulting patterns can pay more attention to important codes and visits for each patient (Section.~\nameref{sec:hen}).
    \item \textbf{Disease detection discriminator for improved pattern classification}. Unlike traditional GAN model that uses a discriminator to classify real or generated data, we construct a disease detector that serves as the discriminator during the adversarial training for identifying candidate positive samples by encouraging a max-margin between the two clusters in the generated complementary samples (Section.~\nameref{sec:detector}). 
    \item \textbf{Complemenrary GAN for boosted pattern augmentation}. \mname uses an adversarial learning mechanism to use ``uncertain'' patients as seeds to generate complementary patient embedding for boosted pattern augmentation (Section.~\nameref{sec:cgan}).
\end{enumerate}

We evaluated \mname on two real-world disease detection tasks (Section.~\nameref{sec:experiments}). The reported results show that for low prevalence inflammatory bowel disease detection, \mname achieved $50.1\%$ relative improvement in PR-AUC and $64.5\%$ in F1 over best baseline medGAN. For rare disease idiopathic pulmonary fibrosis detection, \mname has $41.3\%$ relative improvement in PR-AUC and $39.3\%$ in F1 over best baseline nnPU.
 An additional experiment shows \mname performs well in early disease detection.

\section{Related Work}
\label{sec:relatedwork}
\subsubsection{Data Augmentation via GAN}
GAN consists of a generator that learns to generate new plausible but fake samples and a discriminator that aims to distinguish generated samples from real ones. The two networks are set up in minimax game, where the generator tries to fool the discriminator, and the discriminator aims to discriminate the generated samples~\cite{NIPS2014_5423}. GANs have shown superior performance in image generation, and also demonstrated
 initial success in generating synthetic patient data to address the data limitation. For example, the medGAN model generates synthetic EHR data by combining an autoencoder with GAN~\cite{choi2017generating}. While the ehrGAN model augments patient data in a semi-supervised manner~\cite{che2017boosting}. 
However, existing methods use the generator to fake samples, and once the discriminator is converged, it will not have high confidence in separating samples. 

In this work, we devise a generator which can generate candidate positive and negative samples, and use them to enhance classification performances of the discriminator, rather than distinguishing fake data from real ones.

\subsubsection{Deep Phenotyping}
The availability of massive EHR data enables training of complex deep learning models for accurate predictive health \cite{doi:10.1093/jamia/ocy068}. RETAIN \cite{choi2016retain} uses reverse time attention mechanism to detect influential past visits for heart failure prediction. T-LSTM~\cite{baytas2017patient} handles irregular time intervals in the EHR data. Dipole \cite{ma2017dipole} embeds the visits through a bidirectional GRU for diagnosis prediction. MiME \cite{choi2018mime} leverages auxiliary tasks to improve disease prediction under data insufficiency setting. Despite these achievements, existing works mainly focus on common chronic diseases, while deep learning models for rare disease prediction are lacking.

\subsubsection{Positive-Unlabeled Learning} In PU learning setting, positive samples are determined, while unlabeled samples can either be positive or negative. PU learning has attracted much attention in text classification~\cite{liu2003building}, biomedical informatics~\cite{claesen2015building} and knowledge based completion~\cite{galarraga2015fast}. Two-step approaches ~\cite{zhou2004learning,fung2005text} first extract reliable negative and positive examples and then build the classifier upon them. Direct approaches ~\cite{elkan2008learning,sellamanickam2011pairwise,kiryo2017positive} treat unlabeled examples as negatives examples with class label noise and build the model directly. However, existing methods are not suitable for rare disease detection, as mentioned in the introduction. In this work, we exploit the uncertain samples as seeds to generate candidate positive and negative samples, and build the disease detector by encouraging a max-margin between the generated samples.

\section{Method}
\label{sec:method}

\subsection{Task Description}
\label{sec:task}

\begin{table}[t]
\centering
  \caption{List of basic symbols.
  }
  \label{basic_symbols}
  \begin{tabular}{ll}
    \toprule
    Symbol & Definition and description \\
    \midrule
    $\bm{P}_n$ & The EHR data of the $n$-th patient\\
    $\mathcal{C}_s, \mathcal{C}_p$ & Symptom and procedure code set\\
    $\bm{c}_i^{(t)} \in \{0,1\}^{|\mathcal{C}_\ast|}$ & $i$-th medical code in $t$-th visit\\
    $\mathcal{V}^{(t)}$ & Patient's $t$-th visit\\
    $\textbf{v}^{(t)}$ & Patient's $t$-th visit embedding\\
    $\textbf{s}^{(t)}$ & Annotation of patient's $t$-th visit\\
    $\textbf{u}^{(t)}$ & Weight vector\\
    $\alpha^{(t)}$ & Normalized weight of $t$-th visit\\
    $\mathbf{h}$ & Patient embedding\\
    $\textbf{h}^+\sim p_{\mathbb{R}^+}(\textbf{h})$ & Embeddings of positive patients\\
    $\textbf{h}^-\sim p_{\mathbb{R}^-}(\textbf{h})$ & Embeddings of negative patients\\
    $y\in\{0,1\}$ & Patient's disease label\\
    \midrule
    $F(\bm{P}_n;\theta_e)$ & Hierarchical embedding networks\\
    $D(\mathbf{h};\theta_c)$ & Disease detector\\
    $G(\mathbf{h}^-;\theta_g)$ & Complementary embedding generator\\
    $\hat{y}\in\{0,1\}$ & Disease prediction of patient\\
    $\mathbf{z}\sim p_{\hat{\mathbb{R}}}(\textbf{h})$ & Generated complementary embedding\\
  \bottomrule
\end{tabular}
\end{table}

\begin{definition}[Patient Records]
In longitudinal EHR data, each patient can be represented as a sequence of multivariate observations: $\bm{P}_n=\{\mathcal{V}_n^{(t)}\}^{|\bm{P}_n|}_{t=1}$, 
where $n\in \{1,2,\ldots, N\}$, $N$ is the total number of patients; $|\bm{P}_n|$ is the number of visits of the $n$-th patient. To reduce clutter, we will describe the algorithms for a single patient and drop the subscript $n$ whenever it is unambiguous. For each patient, the visit $\mathcal{V}^{(t)}=\{\bm{c}^{(t)}_1,\dots,\bm{c}^{(t)}_{|\mathcal{V}^{(t)}|}\}$ is a set of several symptom and procedure codes. For simplicity, we use $\bm{c}_i^{(t)}$ to indicate the unified definition for different type of medical codes; $|\mathcal{V}^{(t)}|$ is the number of medical codes. $\bm{c}_i^{(t)} \in \{0,1\}^{|\mathcal{C}_\ast|}$ is a multi-hot vector, where $\mathcal{C}_\ast$ denotes the symptom code set and the procedure code set, and $|\mathcal{C}_\ast|$ is the size of the code set.
\end{definition}

The patient representation/embedding is denoted as $\textbf{h}$. Assume we have $M$ positive embeddings $\textbf{h}^+\sim p_{\mathbb{R}^+}(\textbf{h})$, and $N-M$ negative embeddings $\textbf{h}^-\sim p_{\mathbb{R}^-}(\textbf{h})$, where $\mathbb{R}^+$ and $\mathbb{R}^-$ represents the positive samples' and negative samples' space respectively. $M\ll N$ in extremely imbalanced cases. Table \ref{basic_symbols} lists notations used throughout the paper.

\begin{problem}[Rare Disease Detection]
Given each patient $\bm{P}_n$, we want to learn self-attentive and hierarchical patient embedding net $F: F(\bm{P}_n;\theta_e)\rightarrow \textbf{h}$ to get patient embeddings which becomes the input with a rare disease detection function to determine if the patient has rare disease $D: D(\textbf{h};\theta_c)\rightarrow \hat{y}\in\{0,1\}$. 
\end{problem}


\begin{problem}[Complementary Pattern Augmentation]
Given a set of negative patients' visit embeddings $\mathbf{\textbf{h}^-}$, we want to learn a generator $G$ that can generate complementary embeddings: $G: G(\textbf{h}^-;\theta_g)\rightarrow\mathbf{z}$.
\end{problem}

\subsection{The \mname Framework}
\label{sec:framework}

\begin{figure*}
\centering
\includegraphics[width=.8\textwidth]{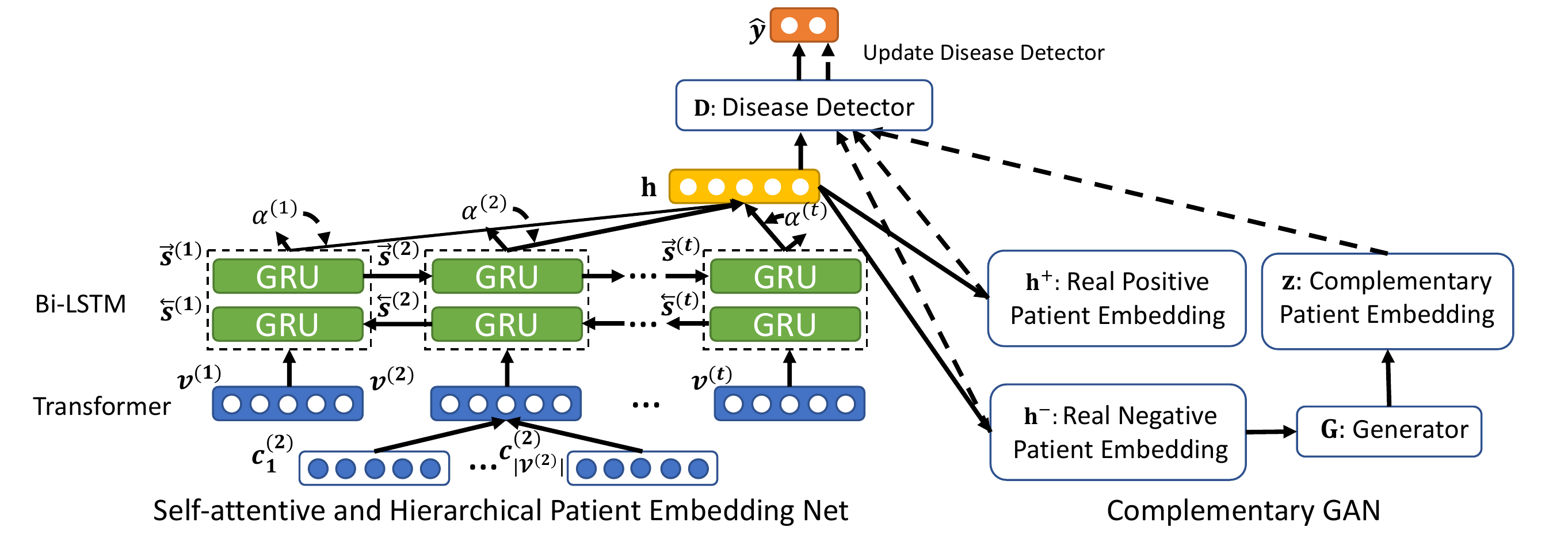}
\vskip -10pt
\caption{Framework overview. It contains three main parts: self-attentive and hierarchical patient embedding net, disease detector and complementary generator. Firstly, we compute the patient's embedding by Eqn. \ref{eqn:transformer}, \ref{eqn:bilstm} and \ref{eqn:att2} through a hierarchical attention mechanism. Then, we use focal loss to train a disease detector. Finally, we set up a complementary generator (Eqn. \ref{eqn:cgan}) to convert negative embeddings to positive with slight modifications, and fine-tune the disease detector with the real and generated embeddings (Eqn. \ref{eqn:finaldetector}). 
}
\vskip -1em
\label{fig:framework}
\end{figure*}

As illustrated in Fig.~\ref{fig:framework}, \mname includes the following components: self-attentive and hierarchical patient embedding net, complementary GAN, and a disease detector component. Next, we will first introduce these modules and then provide details of training and inference of \mname.


\subsubsection{Self-attentive and Hierarchical Patient Embedding Net}
\label{sec:hen}
Motivated by previous work~\cite{ma2017dipole}, we leverage the inherent multilevel structure of EHR data to learn patient embedding. The hierarchical structure of EHR data begins with the patient, followed by visits the patient experiences, then the set of diagnosis codes, procedure, and medication codes recorded for that visit. 

\begin{enumerate}[leftmargin=*,noitemsep]
    \item \textit{Visit Embedding.} Given a patient's visit with codes $\mathcal{V}^{(t)}=\{c^{(t)}_i\}^{|\mathcal{V}^{(t)}|}_{i=1}$, we first embed the codes to vectors via a multi-layer Transformer \cite{vaswani2017attention}, which is a multi-head attention based architecture. It takes the code embeddings as input and derives code embedding $\mathbf{v}^{(t)}$ for a patient at $t$-th visit. As the medical codes in one visit is not ordered, it is proper to use Transformer as it relies entirely on self-attention to compute representations of its input and output without using sequence aligned RNNs or convolution.

Specifically, we remove the position embedding in the Transformer, and get the patient representation of $t$-th visit is computed as follows:
\begin{equation}
    \begin{aligned}
        \textbf{v}^{(t)}=\mathrm{Transformer}(\mathcal{V}^{(t)})
    \end{aligned}
\label{eqn:transformer}
\end{equation}
\item \textit{Patient Embedding.} To capture the patient embedding across multiple hospital visits, we use bidirectional LSTM (Bi-LSTM) as an encoder. An obvious limitation of LSTM is that it can only use previous context. The Bi-LSTM overcomes this gap by incorporating both directions with forward layer and backward layer. To be specific, given visit embedding $\textbf{v}^{(t)}$ of a patient, the patient embedding is computed as below.
\begin{equation}
    \begin{aligned}
        \overrightarrow{\textbf{s}}^{(t)}&=\mathrm{LSTM}(\overrightarrow{\textbf{s}}^{(t-1)}, \textbf{v}^{(t)})\\
        \overleftarrow{\textbf{s}}^{(t)}&=\mathrm{LSTM}(\overleftarrow{\textbf{s}}^{(t+1)}, \textbf{v}^{(t)})
    \end{aligned}
\label{eqn:bilstm}
\end{equation}
We obtain an annotation for visit $\textbf{v}^{(t)}$ by concatenating the forward hidden state $\overrightarrow{\textbf{s}}^{(t)}$ and the backward hidden state $\overleftarrow{\textbf{s}}^{(t)}$, $\textbf{s}^{(t)}=[\overrightarrow{\textbf{s}}^{(t)},\overleftarrow{\textbf{s}}^{(t)}]$, which summarizes the information of the all the visits centered around $\textbf{v}^{(t)}$.

In addition, to find out which visit is important in the whole course, we use self-attention mechanism again to measure the importance of each visit:
\begin{equation}
    \begin{aligned}
        \textbf{u}^{(t)}&=\textbf{u}_s^\top\tanh(\textbf{W}_s\textbf{s}^{(t)}+\textbf{b}_s)\\
        \alpha^{(t)}&=\frac{\exp (\textbf{u}^{(t)})}{\sum_t \exp(\textbf{u}^{(t)})}\\
        \textbf{h}&=\sum_t \alpha^{(t)} \textbf{s}^{(t)}
    \end{aligned}
\label{eqn:att2}
\end{equation}
where $\textbf{u}^{(t)}$ is a hidden representation of $\textbf{s}^{(t)}$ which is obtained by a one-layer MLP and weight vector $\textbf{u}_s$, weight vector $\textbf{u}_s$ is randomly initialized and learned through the training process, $\alpha^{(t)}$ is the normalized weight of the $t$-th visit, and $\mathbf{h}$ is the embedding of a patient which summarizes all the visits and their importance.
\end{enumerate}

The above two-level embedding process of patient's visits is denoted as self-attentive and hierarchical patient embedding net $\mathbf{h}=F(\bm{P}_n;\theta_e)$, where $\theta_e$ represents all the parameters to be learned for simplicity.

\subsubsection{Complementary GAN}
\label{sec:cgan}
As discussed earlier, some negative samples can be converted to positive samples (candidate positive samples) with slight modifications, while others are still negative samples (candidate negative samples). From an adversarial perspective, the disease detector should be ``smart'' enough to classify the candidate positive samples as positive, and others as negative. We denote the candidate positive and negative samples generated as ``complementary samples'', and use them to help enhance the performance of the rare disease detector.

In this section, we introduce a complementary GAN algorithm to generate complementary embedding by using the negative embeddings $\textbf{z}=G(\textbf{h}^-;\theta_g)\sim p_{\hat{\mathbb{R}}}(\textbf{h})$, where $\hat{\mathbb{R}}$ is the distribution space of the complementary samples. The learning purpose of the complementary GAN is to balance the two distribution $p_{\mathbb{R}^+}(\textbf{h})$ and $p_{\hat{\mathbb{R}}}(\textbf{h})$. The generator intends to assign all the generated samples to the positive class with the disease detection $D(\textbf{h};\theta_c)$. Therefore, the loss function is designed as follows:
\begin{equation}
    \begin{aligned}
        \mathcal{L}_g(\theta_g)=&-\lambda\mathbb{E}_{\textbf{h}^-\sim p_{\mathbb{R}^-}}[\log D(G(\textbf{h}^-;\theta_g);\hat{\theta}_c)]\\
        &+||G(\textbf{h}^-;\theta_g)-\textbf{h}^-||_2
    \end{aligned}
\label{eqn:cgan}
\end{equation}
where the first term measures the difference between the output probability and the positive distribution, and the second term measures the change between the original negative samples and the converted samples. $\lambda$ is a trade-off parameter that controls the relative importance of the two terms. We tried $\lambda=\{0.01, 0.05, 0.1, 0.5, 1\}$ and $\lambda=0.05$ works best.

\subsubsection{Disease Detector}
\label{sec:detector}
The disease detector can classify both generated and original embedding to learn important codes/visits. It is built on top of the hierarchical embedding, taking the patient representation $\textbf{h}$ as input. We denote the disease detector as $D(\cdot;\theta_c)$, where $\theta_c$ represents the parameters to be learned. The output of the disease detector is the probability of this patient being positive.

The goal of the disease detector if to determine whether a patient has the disease or not. In rare disease detection problem, the discriminator evaluate $10^4-10^5$ patients but only a few have the target rare disease. Such a data imbalance issue causes two learning problems: (1) the training is insufficient as the easy negatives do not contain much information; (2) the easy negatives might degenerate the model. To efficiently train on all examples, we employ focal loss \cite{lin2017focal}:
\begin{align}\label{eqn:focal}    &\mathcal{L}_c(\theta_e,\theta_c)=-\mathbb{E}_{\textbf{h}^+\sim p_{\mathbb{R}^+}}[\alpha (1-D(\textbf{h};\theta_c))^\gamma \log D(\textbf{h};\theta_c)] \nonumber \\
      & -\mathbb{E}_{\textbf{h}^-\sim p_{\mathbb{R}^-}}[(1-\alpha)D(\textbf{h};\theta_c)^\gamma \log (1-D(\textbf{h};\theta_c))]
\end{align}
where $\gamma$ is a focusing parameter, which focuses more on hard and easily mis-classified examples, and $\alpha$ is the weight assigned to the rare class. $\gamma=2$ and $\alpha=0.25$ works best based on the rule of thumb~\cite{lin2017focal}. 

The disease detector is trained as follows. First, the disease detector works with the embedding net to get patient representation $\textbf{h}$. Second, with all the negative samples that are converted as positive through the generator, we want the detector to distinguish the candidate positive samples from the candidate negative samples, through maximizing the distance between two classes. Thus, we use the conditional entropy of labelings \cite{dai2010minimum,deng2017disguise} to maximize the margin:
\begin{equation}
    \begin{aligned}
        H(\theta_c)=&-D(\textbf{z};\theta_c)\log(D(\textbf{z};\theta_c))\\
        &-(1-D(\textbf{z};\theta_c))\log(1-D(\textbf{z};\theta_c))
    \end{aligned}
\label{eqn:entropy}
\end{equation}

So we combine this goal with focal loss to form the final loss of the detector to further fine-tuning the parameter $\theta_c$:
\vspace{-2pt}
\begin{equation}
    \begin{aligned}
        \mathcal{L}_s(\theta_c)=&\mathcal{L}_c(\hat{\theta}_e,\theta_c)+\eta H(\theta_c)
    \end{aligned}
\label{eqn:finaldetector}
\end{equation}
where $\eta$ is a weighting factor. We set $\eta=0.05$.

\subsection{Training and Inference with \mname}
\label{sec:training}

During the training stage, in the first stage, the self-attentive and hierarchical patient embedding net $F(\cdot; \theta_e)$ work with disease detector $D(\cdot;\theta_c)$ to minimize the detection loss $\mathcal{L}_c(\theta_e,\theta_c)$, so as to provide insight into which codes and visits contribute to the rare disease. In the second stage, the generator $G(\cdot;\theta_g)$ tries to fool the detector $D(\cdot;\theta_c)$ by minimizing the loss $\mathcal{L}_g(\theta_g)$. Also by minimizing the discrimination loss $\mathcal{L}_s(\theta_c)$, the detector $D(\cdot;\theta_c)$ not only tries to discriminate positive embeddings from negative embeddings, but also maximize the margin between the two clusters of generated samples. The detailed training steps are summarized in Algorithm \ref{alg:1}.

\begin{algorithm}[!htb]
\caption{\mname for Rare Disease Detection.}
\label{alg:1}
\KwIn{Training set, training epochs for self-attentive and hierarchical patient embedding net $N_e$ and complementary GAN $N_g$} 
\KwOut{Well-trained rare disease detector and complementary patient data embedding $\textbf{z}$}
\For{$i=1,\dots,N_e$} 
{
Minimize the detection loss $\mathcal{L}_c(\theta_e,\theta_c)$ in Eq. \ref{eqn:focal}\;
}
\ForEach{patient in dataset}
{
Compute patient's embedding $\textbf{h}$ by Eq. \ref{eqn:transformer}, \ref{eqn:bilstm} and \ref{eqn:att2}\;
}
\For{$i=1,\dots,N_g$} 
{
Minimize the generation loss $\mathcal{L}_g(\theta_g)$ in Eq. \ref{eqn:cgan} with negative patient's embedding $\textbf{h}^-$\;
Compute the generated samples $\textbf{z}$\;
Feed the detector with both the original patient's embedding $\textbf{h}$ and generated sample $\textbf{z}$\;
Minimize the final detection loss $\mathcal{L}_s(\theta_c)$ in Eq. \ref{eqn:finaldetector}\;
}
\end{algorithm}

\section{Experiments}
\label{sec:experiments}

\subsection{Experimental Setup}

\subsubsection{Data}

We leverage data from IQVIA longitudinal prescription (Rx) and medical claims (Dx) databases, which include hundreds of millions patients’ clinical records. In our study, we focus on one rare disease and one low prevalence disease.
\begin{enumerate}
    \item \textbf{Idiopathic Pulmonary Fibrosis (IPF)} is a pulmonary disease that is characterized by the formation of scar tissue within the lungs in the absence of any known provocation~\cite{Meltzer2008}. IPF is a rare disease which affects approximately 5 million persons worldwide, with prevalence rate at $0.04\%$
    \item \textbf{Inflammatory Bowel Disease (IBD)} is a broad term that describes conditions characterized by chronic inflammation of the gastrointestinal tract. The two most common inflammatory bowel diseases are ulcerative colitis and Crohn's disease. IBD has a low prevalence. Overall, the prevalence of IBD is 439 cases per 100,000 persons.
\end{enumerate}
For both datasets, we extracted patient records, including medical/diagnoses/procedure codes at visit level from January 2010 to April 2019, which include total 168,514 distinct medical codes (ICD-9). Each visit contains patient id,  time of the visit, one symptom code and one diagnose/procedure. Data statistics are provided in Table \ref{tab:datasets}.

\begin{table}[!htb]
\caption{Statistics of datasets. The disease prevalance rates are the same as case/control ratio in test set.}
\label{tab:datasets}
\centering
\resizebox{0.8\columnwidth}{!}{
\begin{tabular}{lcc}
\toprule
 & IPF &  IBD \\ \hline
Category & rare &  low prevalence  \\
Prevalence & 0.04\% &  0.44\% \\
Positive & 9,996 &  1,405  \\
Negative & 24,757,572 &  108,047 \\
Ave. \# of visit & 597.36 &  798.25 \\ \bottomrule
\end{tabular}
}
\vskip -10pt
\end{table}

\subsubsection{Baselines}

We consider the following baseline methods:
\begin{itemize}
    \item \textbf{LR}: We first embed each code into a vector, then concatenate all the vectors together, and feed it to the model.
    \item \textbf{PU-SVM} \cite{elkan2008learning}: PU-SVM is a well-known PU learning model. It labels positive training examples at random. The data are processed similarly to LR.
    \item \textbf{nnPU} \cite{kiryo2017positive}: A PU learning model that is more robust against overfitting, and allows for using deep neural networks given limited positive data. The data are processed similarly to LR.
    \item \textbf{RNN}: We feed the code embeddings to a fully-connected RNN. The output generated by the RNN is directly used to predict the rare disease. 
    \item \textbf{T-LSTM}~\cite{baytas2017patient}: T-LSTM handles time irregularity. Similar to RNN, we feed the embeddings to T-LSTM model, and use the output to predict the disease.
    \item \textbf{SMOTE} \cite{chawla2002smote}: SMOTE is an oversampling method which balances class distribution by randomly increasing minority class examples by replicating them. We use LR and RNN to generate embeddings, and then use SMOTE to augment the positive class.
    \item \textbf{RETAIN} \cite{choi2016retain}: RETAIN is a two-level attention-based neural model which detects influential past visits and significant clinical variables within those visits such as key diagnoses. It receives the EHR data in a reverse time to mimic the practical clinical course.
    \item \textbf{medGAN} \cite{choi2017generating}: medGAN generates synthetic EHR data. We first generate both positive and negative EHR data to balance the class distribution, and feed the output of the encoder into an MLP with cross-entropy loss to predict disease.
    \item \textbf{SSL GAN} \cite{yu2019rare}: SSL GAN can augment positive embeddings, and facilitate rare disease detection by leveraging both positive and negative samples.
    \item \textbf{Dipole} \cite{ma2017dipole}: Dipole employs bidirectional recurrent neural networks to embed the EHR data. It introduces the attention mechanism to measure the relationships of different visits for the diagnosis prediction.
\end{itemize}

\subsubsection{Metrics}
To measure the prediction performance, we used the following three metrics:
\begin{enumerate}
    \item \textbf{Area Under the Precision-Recall Curve}~(PR-AUC):
    \vspace{-4pt}
    \begin{equation*}
        \text{PR-AUC} = \sum_{k = 1}^{n} \mathrm{Prec}(k) \Delta \mathrm{Rec}(k), 
    \end{equation*}
    where $k$ is the $k$-th precision and recall operating point ($\mathrm{Prec}(k), \mathrm{Rec}(k)$).
  
    \item \textbf{F1 Score}: $ \mathrm{F1~Score} = 2\cdot(\mathrm{Prec}\cdot\mathrm{Rec})/(\mathrm{Prec}+\mathrm{Rec})$, where $\mathrm{Prec}$ is precision and $\mathrm{Rec}$ is recall.
 \item \textbf{Cohen's Kappa}: $\kappa=(p_o-p_e)/(1-p_e)$, where $p_o$ is the observed agreement (identical to accuracy), and $p_e$ is the expected agreement, which is probabilities of randomly seeing each category.
\end{enumerate}

\subsubsection{Implementation Details}

We implement all models with Keras~\footnote{https://github.com/cuilimeng/CONAN}. We sample two imbalanced training sets for each dataset, with a ratio of $10\%$ and $1\%$ for positive samples. For the testing set, we extract the data using the actual disease prevalence rate shown in Table \ref{tab:datasets}. We set $128$ for dimensions of patient embedding. For the complementary GAN, the disease detector serves as the discriminator, and the complementary generator has two hidden layers with 128 dimensions. The output layer of the generator has the same dimension as the patient embedding. The training epoch of complementary GAN is 1000.
For all models, we use RMSProp \cite{hinton2012neural} with a mini-batch of $512$ patients, and the training epoch is $30$. In order to have a fair comparison, we use focal loss (with $\gamma=2$ and $\alpha=0.25$) and set the output dimension as $128$ for all models. The vocabulary size is consistent with ICD-9 diagnosis codes. The sequence length is chosen according to the average number of visit per patient in Table \ref{tab:datasets}. For RNN and LSTM, the hidden dimensions of the embedding layer are set as $128$. For other methods, we follow the network architectures in the papers. All methods are trained on an Ubuntu 16.04 with 128GB memory and Nvidia Tesla P100 GPU.

\begin{table*}[!htb]
\caption{Performance Comparison on \textbf{ IPF (rare disease, prevalence rate $0.04\%$)} and \textbf{ IBD (low prevalence disease, prevalence rate $0.44\%$ } ) datasets. \mname outperforms all state-of-the-art baselines including GAN based and PU learning baselines.}
\label{tab:results}
\centering
\resizebox{\textwidth}{!}{
\begin{tabular}{lccccccccccccc}
\toprule
Dataset & Metric & LR & PU-SVM & nnPU & RNN & T-LSTM & SMOTE$_{LR}$ & SMOTE$_{RNN}$ & RETAIN & Dipole & SSL GAN & medGAN & \mname \\ \hline
\multirow{3}{*}{IBD} & PR-AUC & 0.2765 & 0.5321 & 0.5682 & 0.4373 & 0.2241 & 0.3464 & 0.4471 & 0.3135 & 0.5417 & 0.6072 & 0.6385 & \textbf{0.9584} \\
 & F1 Score & 0.3651 & 0.4982 & 0.4392 & 0.4332 & 0.3016 & 0.4341 & 0.4642 & 0.3594 & 0.5528 & 0.5416 & 0.5834 & \textbf{0.9601} \\
 & Cohen's Kappa & 0.3249 & 0.5123 & 0.4624 & 0.4440 & 0.2886 & 0.3451 & 0.4895 & 0.3106 & 0.5904 & 0.5453 & 0.4851 & \textbf{0.9595} \\ \hline
\multirow{3}{*}{IPF} & PR-AUC & 0.0798 & 0.1141 & 0.1578 & 0.0090 & 0.0084 & 0.0406 & 0.0187 & 0.1016 & 0.1183 & 0.0206 & 0.0954 & \textbf{0.2229} \\
 & F1 Score & 0.1529 & 0.0915 & 0.1682 & 0.0169 & 0.0211 & 0.0673 & 0.0293 & 0.1345 & 0.0969 & 0.0272 & 0.0729 & \textbf{0.2343} \\
 & Cohen's Kappa & 0.1369 & 0.0835 & 0.1397 & 0.0261 & 0.0752 & 0.0208 & 0.0429 & 0.1470 & 0.1060 & 0.0372 & 0.0612 & \textbf{0.2339} \\ \bottomrule
\end{tabular}
}
\vskip -10pt
\end{table*}



\subsection{Results}

\subsubsection{Performance Comparison}
For the training data, we tried positive sample ratios, $10\%$ and $1\%$, and report better results of two ratios on IBD and IPF dataset in Table \ref{tab:results}. The best results are presented in bold figures. For IBD, the methods perform better by using $10\%$ positive samples. For IPF, the methods perform better with $1\%$ positive samples. We assume that the method performs best when the ratio of positive in training data is close to the ratio in test data, which is the actual prevalence rate of that disease. We also tried $0.1\%$ and $50\%$ positive on both datasets but did not get satisfying results. For $0.1\%$ positive, all the samples are classified into the negative class regardless of the combinations of hyperparameters. The medical knowledge about negative patients suggests us that they can not be strictly counted as a class, so they contribute little to the identification of positive patients. For $50\%$ positive, the false positive rate is high on the test data, which indicates that this training strategy is not practical for the real-world problem.

From Table \ref{tab:results} we can observe that the performance of LR, RNN and T-LSTM are less satisfactory, due to the complexity of the disease progression during long clinical courses.

Among the hierarchical embedding methods, RETAIN and Dipole are two hierarchical embedding methods with an attention mechanism. We can observe that Dipole performs better than RETAIN. For RETAIN, it models the EHR data in a reverse time to mimic the practical clinical course. The recent visits receive greater attention than previous visits. It is not very practical for the rare disease. Due to the complexity of the rare disease, the diagnoses/symptoms of similar diseases may intertwine together during the disease progression.

Regarding PU learning methods, including PU-SVM and nnPU, they are designed to exploit the reliable negative cases in the unlabeled data. However, these methods may not be suitable for the rare disease detection problem. As the main challenge in rare disease detection is to distinguish the patients with the rare disease (e.g., IPF) to patients with similar diseases (e.g., hypersensitivity pneumonitis), but the detected reliable negative cases are healthy people or patients with irrelevant diseases, which contribute little to the detection.

Oversampling (SMOTE$_{LR}$ and SMOTE$_{RNN}$) and generative (medGAN and SSL GAN) models perform better than simple sequential models in most cases, which shows the effectiveness of the data augmentation.

\begin{table}[!htb]
\caption{Abalation study of \mname demonstrated the advantage of complementary pattern augmentation.}
\label{tab:ablation}
\centering
\resizebox{\columnwidth}{!}{
\begin{tabular}{lcccc}
\toprule
Dataset & Metric & w/o GAN & w GAN &  \mname \\ \hline
\multirow{3}{*}{IBD} & PR AUC & 0.8097 & 0.9323 & 0.9584 \\
 & F1 Score & 0.8386 & 0.9566 & 0.9601 \\
 & Cohen's Kappa & 0.8590 & 0.9560 & 0.9595 \\ \hline
\multirow{3}{*}{IPF} & PR AUC & 0.1796 & 0.2023 & 0.2229 \\
 & F1 Score & 0.1119 & 0.1768 & 0.2343 \\
 & Cohen's Kappa & 0.0767 & 0.1762 & 0.2339 \\ \bottomrule
\end{tabular}
}
\end{table}

\subsubsection{Ablation Study}
We conduct an ablation study to understand the contribution of each component in \mname. We remove/change the GAN module as below. The parameters in all the variants are determined with cross-validation, and the best performances are reported in Table \ref{tab:ablation}.

\begin{itemize}
    \item w/o GAN: w/o GAN is a variant of \mname, which only contains the hierarchical patient embedding net.
    \item w GAN: w GAN is a variant of \mname, which replaces the complementary GAN with the regular GAN, which is used to augment the positive embeddings.
    \item w cGAN: this method is \mname, which incorporates the complementary GAN.
\end{itemize}

The results indicate that, when we solely use the self-attentive and hierarchical patient embedding net, the performances are largely reduced. It suggests the necessity of data augmentation. When we use the regular GAN to augment the data, performance are improved but still lower than \mname. By augmenting the positive samples, the disease detector will have more confidence in detecting the positive samples. In other words, it detects more samples as positive, which yields a much higher false positive rate but doesn't decrease the false negative rate much as the ratio of positive samples is extremely low.

Through the ablation study of \mname, we conclude that (1) data augmentation can contribute to the low prevalence rate disease detection performance; (2) complementary GAN is necessary for rare disease detection.

\subsubsection{Early Disease Prediction}
We test how \mname performs on early disease prediction for unseen patients. We select the first $x\%$ of visits in each patient's records for testing, where $x$ is varied as $\{100, 50, 20 \}$. Table \ref{tab:earlydetection} shows comparison results in terms of the PR AUC, F1 Score and Cohen's Kappa of early disease prediction. The \mname achieves a satisfactory performance when $x=50$ and $x=20$, and it is still competitive compared with the baselines. It indicates that once we train the model, for unseen patients, we can predict their conditions at an early stage.

\begin{table}[!htb]
\caption{The results on early prediction indicated that we can use \mname to predict patients' conditions at an early stage.
}
\label{tab:earlydetection}
\centering
\resizebox{\columnwidth}{!}{
\begin{tabular}{lccccc}
\toprule
\multirow{2}{*}{Dataset} & \multirow{2}{*}{Metric} & \multicolumn{3}{c}{\% Visits in Test Data} \\
 &  & 100\% & 50\% & 20\% \\ \hline
\multirow{3}{*}{IBD} & PR AUC & 0.9584 & 0.9474 & 0.7313 \\
 & F1 Score & 0.9601 & 0.9531 & 0.7993 \\
 & Cohen's Kappa & 0.9595 & 0.9473 & 0.7629 \\ \hline
\multirow{3}{*}{IPF} & PR AUC & 0.2229 & 0.2105 & 0.0843 \\
 & F1 Score & 0.2343 & 0.2262 & 0.1024 \\
 & Cohen's Kappa & 0.2119 & 0.2056 & 0.0758 \\ \bottomrule
\end{tabular}
}
\vskip -10pt
\end{table}


\subsubsection{Visualize Generated Embedding}
\label{sec:visualization}

\begin{figure}[!htb]
    \centering
    \begin{subfigure}[t]{0.48\columnwidth}
        \centering
        \includegraphics[width=\columnwidth]{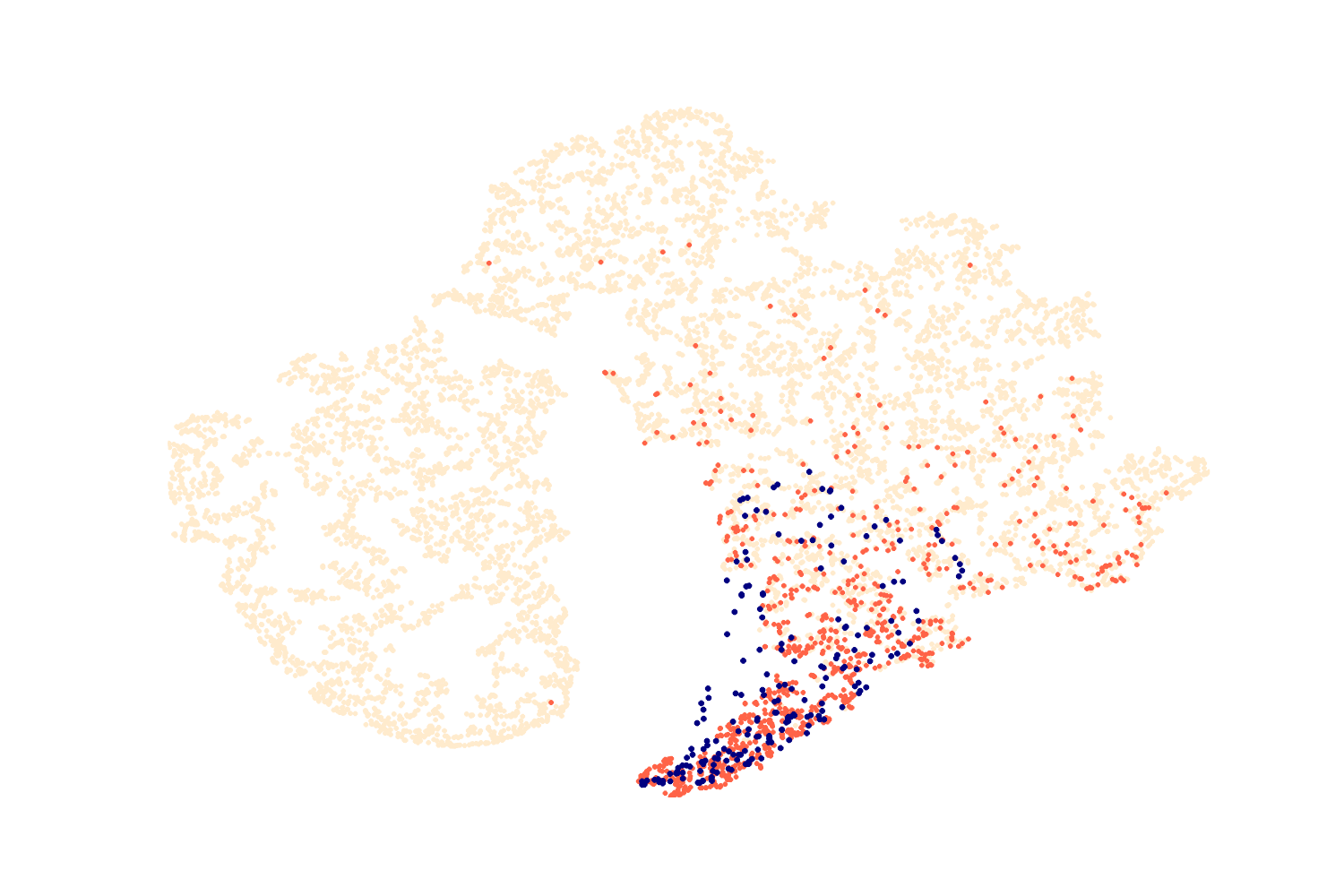}
        \caption{Regular GAN}
        \label{fig:emb1}
    \end{subfigure}%
    \hspace{0.01\linewidth}
    \begin{subfigure}[t]{0.48\columnwidth}
        \centering
        \includegraphics[width=\columnwidth]{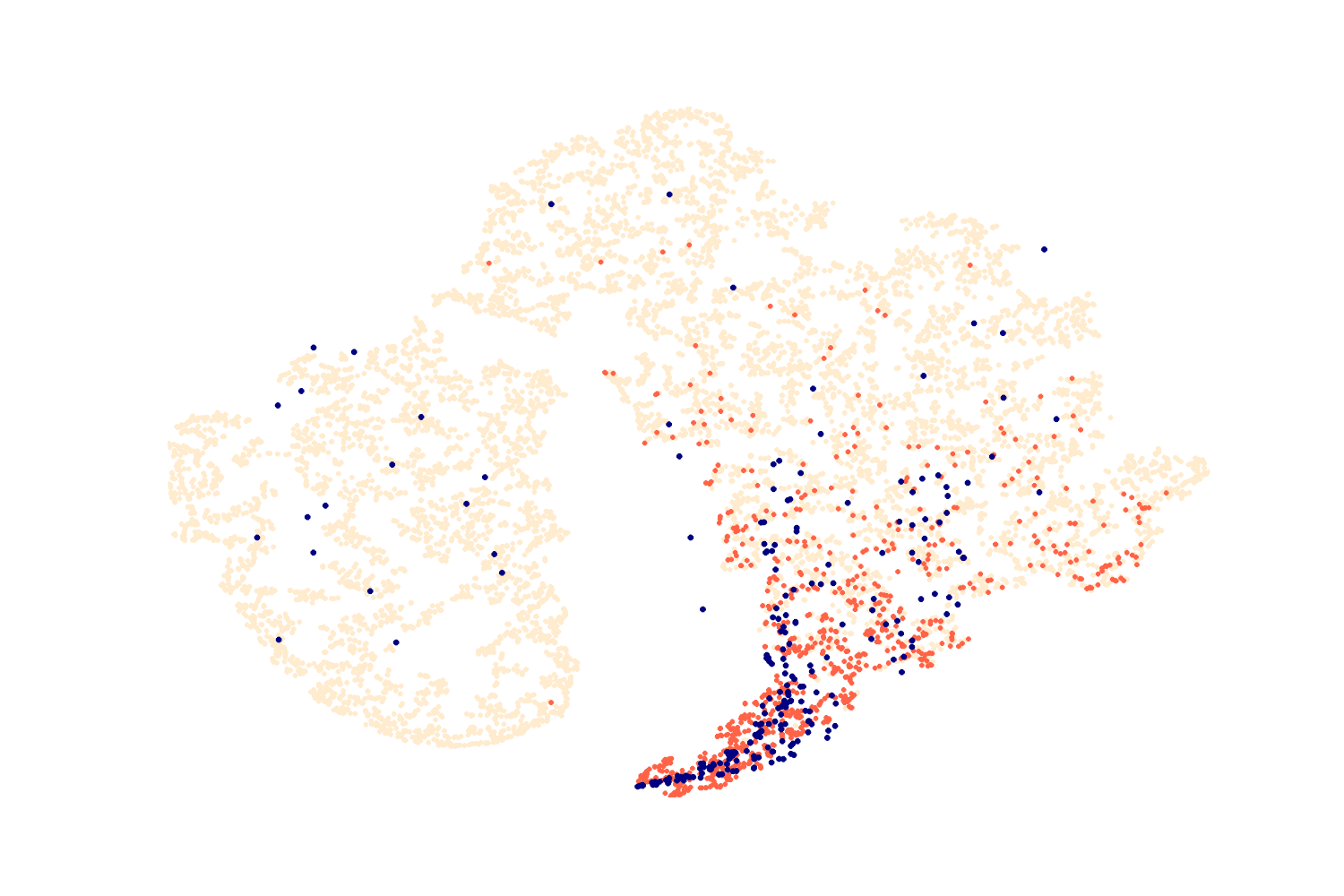}
        \caption{medGAN}
        \label{fig:emb4}
    \end{subfigure}
    \vfill
    \begin{subfigure}[t]{0.48\columnwidth}
        \centering
        \includegraphics[width=\columnwidth]{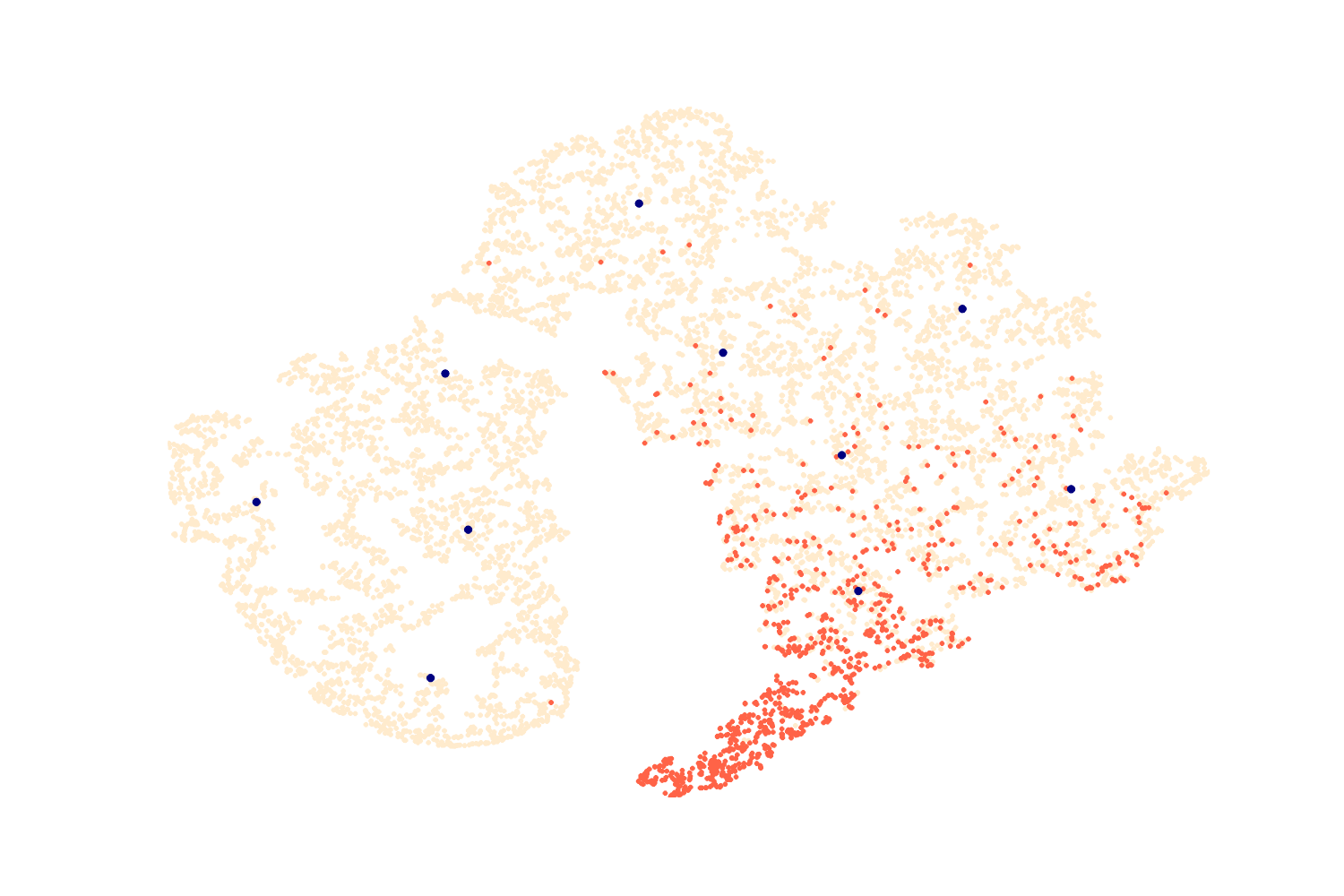}
        \caption{PU Learning}
        \label{fig:emb3}
    \end{subfigure}
    \hspace{0.01\linewidth}
    \begin{subfigure}[t]{0.48\columnwidth}
        \centering
        \includegraphics[width=\columnwidth]{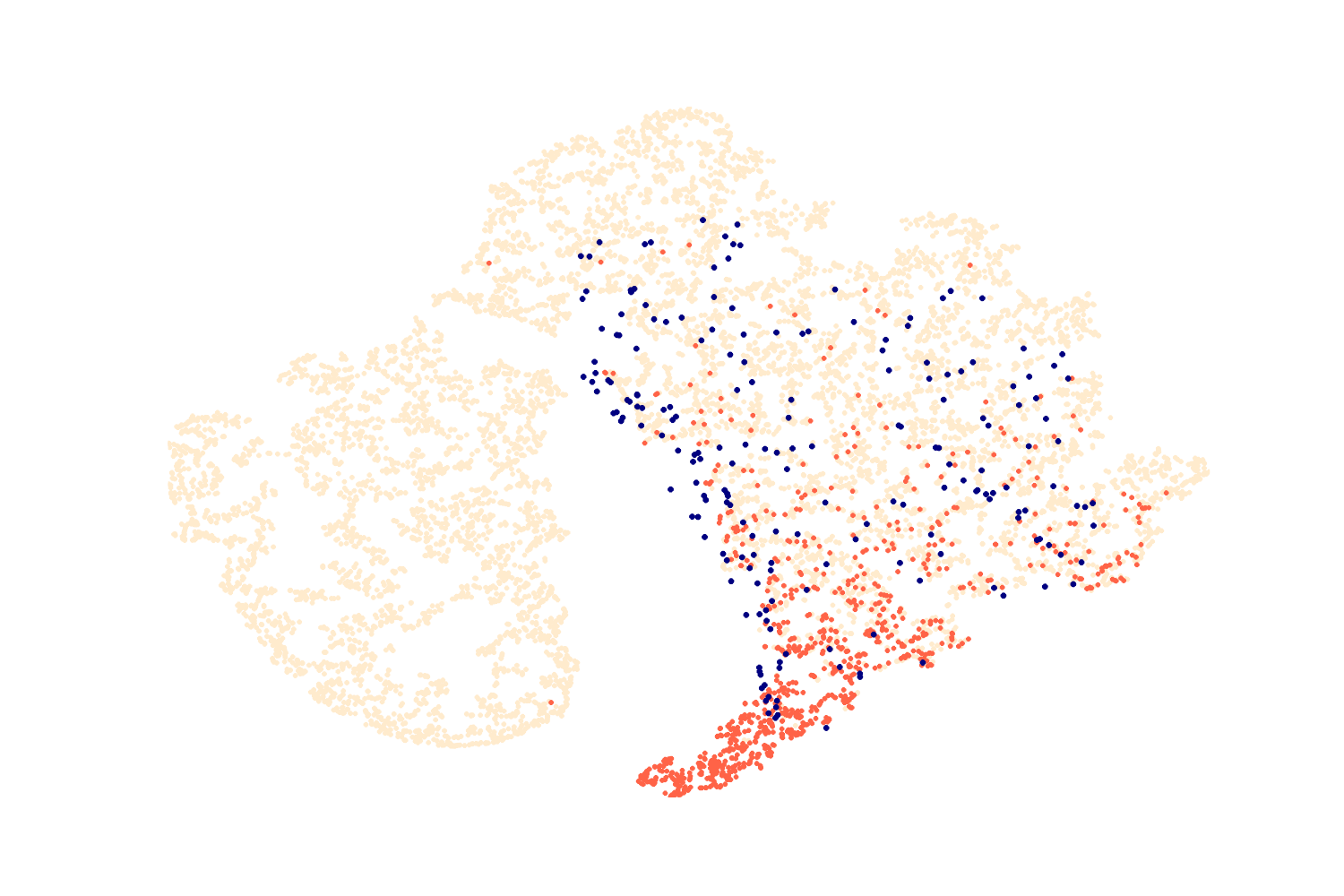}
        \caption{\mname}
        \label{fig:emb2}
    \end{subfigure}
    \caption{2D visualization of patient embeddings of IPF dataset: real negative (almond), real positive (red), and generated (navy). The complementary samples generated by \mname lie in between of real positive and negative samples, which can help the detector update its hyperplane by encouraging a max margin between the generated samples. } 
    \vskip -10pt
    \label{fig:emb}
\end{figure}

We project the three types of patient's embeddings of IPF dataset, including real positive, real negative and generated, to a two-dimensional space by t-SNE \cite{maaten2008visualizing} and show the projection in Fig.~\ref{fig:emb}. The red dots indicate the positive embeddings. The almond dots indicate the negative embeddings. The navy dots in four subfigures indicate the samples generated by regular GAN, medGAN, a PU learning method \cite{liu2003building}, and \mname respectively. The generated samples of the four methods are distributed differently. The samples generated by regular GAN have the same distribution with the real positive samples. As a result, it would give the detector more confidence to classify the ``borderline'' patients into positive, which may yield a high false positive rate. We use medGAN to generate both positive and negative cases to balance the overall class distribution, but the same issue as regular GAN is still unsolved. The reliable negative samples generated by the PU learning method span the negative samples' space, which contribute little to the rare disease detection. 
The complementary samples generated by \mname lie in between of real positive and negative samples, which can help the disease detector update its hyperplane by encouraging a max-margin between the generated samples.

\section{Conclusion}

In this paper, we proposed \mname, a pattern augmentation method for rare and low prevalence disease detection. \mname uses the embedding of negative samples as seeds to generate complementary patterns with a complementary GAN. The generator can convert the negative embedding to fool the discriminator, while the disease detector serves as the discriminator to distinguish the positive and negative samples by maximizing a margin between the generated samples. After the training, the discriminator can be used for detecting positive patients. Experiments on real-world datasets demonstrated the strong performance of \mname. 

The \mname method can also be extended to other application domains for classification problems with imbalanced data, such as fraud detection and recommendation. Besides, there are several interesting future directions that need investigations. First, we can incorporate the data from similar diseases to guide the generation process and obtain more distinguishable patient embeddings. Second, other data sources, such as the doctor notes can be considered for better embedding. Third, time intervals between visits can be considered for modeling the progression of rare disease.

\section*{Acknowledgement}
This work was in part supported by the National Science Foundation award IIS-1418511, CCF-1533768 and IIS-1838042, the National Institute of Health award NIH R01 1R01NS107291-01 and R56HL138415.

\bibliographystyle{aaai}
\bibliography{sample}

\clearpage

\appendix

\section{Experiment on a Common Disease}

We also test the performance of \mname on a common disease NASH (Nonalcoholic Steatohepatitis), with prevalence rate at $7.58\%$. It is a liver disease that resembles alcoholic liver disease, but occurs in people who drink little or no alcohol. The major feature in NASH is fat in the liver, along with inflammation and damage. Although the prevalence rate of this disease is relatively high, but it still occurs an imbalanced learning problem when using the data for disease detection. Data statistics are provided in Table \ref{tab:datasets_nash}.

The results are shown in Table \ref{tab:results_nash}. Our method still outperforms the state-of-the-art results $2.56\%$ on PR AUC and $10.10\%$, which indicates the feasibility of our method. Unlike rare disease, common disease can often be confirmed by several symptoms and diagnoses in one visit without misdiagnosis. This explains why LR can achieve satisfactory results. As for \mname, we remove the position embedding in the original Transformer while encoding each visit, which does not enforce the symptoms/diagnoses to be sequential in each visit.

\begin{table}[!htp]
\caption{Statistics of NASH dataset. The disease prevalance rates are the same as case/control ratio in test set.}
\vskip 0.5em
\label{tab:datasets_nash}
\centering
\resizebox{0.5\columnwidth}{!}{
\begin{tabular}{lccc}
\toprule
  & NASH  \\ \hline
Category  & common   \\
Prevalence  & 7.58\% \\
Positive  & 4,380  \\
Negative & 57,769  \\
Ave. \# of visit  & 241.47  \\ \bottomrule
\end{tabular}
}
\end{table}

\begin{table}[ht]
\caption{Performance Comparison on NASH dataset.}
\label{tab:results_nash}
\centering
\begin{tabular}{lccc}
\toprule
Method & PR-AUC & F1 Score & Cohen's Kappa\\\hline
LR & 0.5930 & 0.5471 & 0.5380 \\
PU-SVM & 0.3971 & 0.3582 & 0.3619 \\
nnPU & 0.4484 & 0.3824 & 0.4139 \\
RNN & 0.1841 & 0.3739 & 0.4313 \\
T-LSTM & 0.3446 & 0.1783 & 0.2448 \\
SMOTE$_{LR}$ & 0.6123 & 0.5732 & 0.5842 \\
SMOTE$_{RNN}$ & 0.2091 & 0.4195 & 0.4485 \\
RETAIN & 0.5641 & 0.4627 & 0.4506 \\
Dipole & 0.5929 & 0.5221 & 0.5477 \\
SSL GAN & 0.5397 & 0.5340 & 0.4883 \\
medGAN & 0.4825 & 0.3986 & 0.3822 \\
\mname & \textbf{0.6186} & \textbf{0.6481} & \textbf{0.5958} \\
\bottomrule
\end{tabular}
\end{table}


\section{Data Visualization}

We compared the generated patterns of the proposed framework, \mname, with 3 baseline methods discussed in Section.~\nameref{sec:visualization}, including regular GAN, medGAN and PU learning method. We first use the self-attentive and hierarchical patient embedding net to embed the positive and negative samples into vectors with $128$ dimensions, then we use t-SNE to project the embeddings to a two-dimensional space. Finally, we apply the methods on the embeddings to maintain the consistency and plot the generated samples. Specifically, we apply each method as follows:

\begin{itemize}
    \item \textbf{Regular GAN}: We generate more positive samples.
    \item \textbf{medGAN}: We generate both positive and negative samples to balance the overall class distribution.
    \item \textbf{PU Learning}: We use Biased SVM \cite{liu2003building} to generate the reliable negative samples.
    \item \textbf{\mname}: We use the complementary GAN in \mname to generate the complementary embeddings.
\end{itemize}

\section{Reproducibility}
All codes that we have implemented and adapted, including \mname and baselines, are publicly available online\footnote{https://github.com/cuilimeng/CONAN}. For the dataset, we create a sample of the dataset to elaborate the format of the data used. As for baseline methods, we implemented LP, RNN and SSL GAN by ourselves on Keras. For other baselines we used the codes provided by the authors.

\end{document}